\documentclass[letterpaper]{article} 
\usepackage{aaai2026}
\nocopyright
\usepackage{times}  
\usepackage{helvet}  
\usepackage{courier}  
\usepackage[hyphens]{url}  
\usepackage{graphicx} 
\usepackage{natbib}  
\usepackage{caption} 
\usepackage{booktabs}
\usepackage{multirow}
\usepackage{subcaption}
\usepackage{pdfpages}
\usepackage{algorithm}
\usepackage{algorithmic}
\usepackage{booktabs}

\usepackage{newfloat}
\usepackage{listings}
\usepackage{subcaption}
\usepackage{graphicx}   
\usepackage{graphicx}
\usepackage{xcolor}
\DeclareCaptionStyle{ruled}{labelfont=normalfont,labelsep=colon,strut=off} 
\floatstyle{ruled}
\newfloat{listing}{tb}{lst}{}
\floatname{listing}{Listing}
\usepackage{xspace}
\newcommand{\SchemaCoder}{\textsc{SchemaCoder\xspace}}
\newcommand{\GPTFo}{\textsc{GPT-4o\xspace}}

\newcommand{\QTree}{\textsc{Q-Tree-PR\xspace}}

\title{\SchemaCoder: A Q-Tree Driven Automatic Schema Extraction Coder of Diverse Logs Using LLMs with Feedback Evolution and Boosting}
\title{\SchemaCoder: Automatic Log Schema Extraction \\ Coder with Residual Q-Tree Boosting Mechanism}
\title{\SchemaCoder: Automatic Log Schema Extraction \\ Coder with Residual Q-Tree Boosting}
\author{
  Lily Jiaxin Wan\textsuperscript{\rm 1}\textsuperscript{\rm 2}\textsuperscript{\rm *},
  Chia-Tung Ho\textsuperscript{\rm 1}\textsuperscript{\rm *},
  Rongjian Liang\textsuperscript{\rm 1},
  Cunxi Yu\textsuperscript{\rm 1}\textsuperscript{\rm 3},
  Deming Chen\textsuperscript{\rm 2},
  Haoxing Ren\textsuperscript{\rm 1}
}
\affiliations{
  \textsuperscript{\rm 1}NVIDIA, USA\\
  \textsuperscript{\rm 2}Univerisity of Illinois, Champaign, USA\\
  \textsuperscript{\rm 3}University of Maryland, College Park, USA\\
  \textsuperscript{\rm *}Corresponding authors: liwan@nvidia.com, chiatungh@nvidia.com
}

\usepackage{bibentry}

\begin{document}

\maketitle

\begin{abstract}
Log schema extraction is the process of deriving human-readable templates from massive volumes of log data, which is essential yet notoriously labor-intensive.
Recent studies have attempted to streamline this task by leveraging Large Language Models (LLMs) for automated schema extraction.
However, existing methods invariably rely on predefined regular expressions, necessitating human domain expertise and severely limiting productivity gains.
To fundamentally address this limitation, we introduce \SchemaCoder, the first fully automated schema extraction framework applicable to a wide range of log file formats without requiring human customization within the flow.
At its core, \SchemaCoder~features a novel Residual Question-Tree (Q-Tree) Boosting mechanism that iteratively refines schema extraction through targeted, adaptive queries driven by LLMs.
Particularly, our method partitions logs into semantic chunks via context-bounded segmentation, selects representative patterns using embedding-based sampling, and generates schema code through hierarchical Q-Tree-driven LLM queries, iteratively refined by our textual-residual evolutionary optimizer and residual boosting.
Experimental validation demonstrates \SchemaCoder's superiority on the widely-used LogHub-2.0 benchmark, achieving an average improvement of 21.3\% over state-of-the-arts.

\end{abstract}


\section{Introduction}

\begin{figure*}[t]
  \centering
  \begin{subfigure}[t]{0.495\linewidth}
    \includegraphics[width=\linewidth]{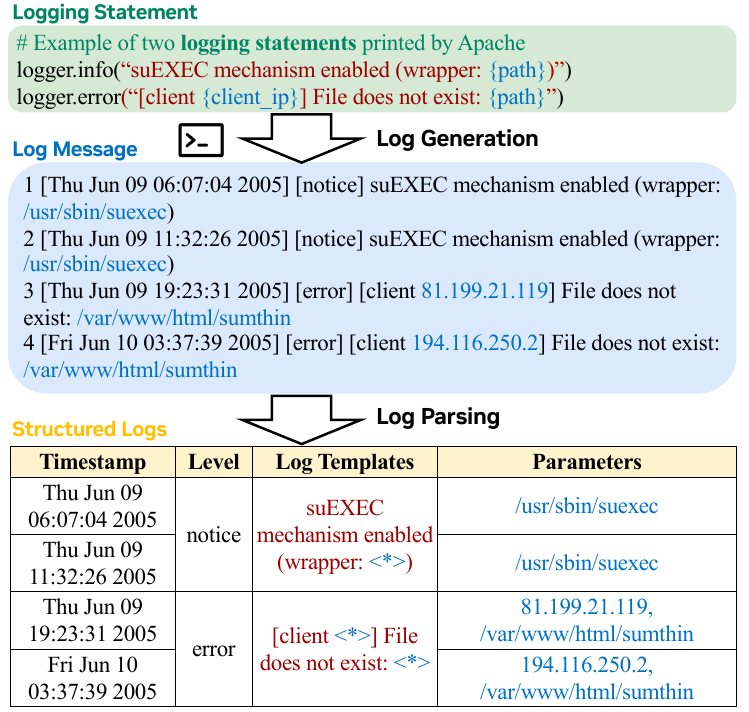}
    \caption{LogHub-2.0~\cite{jiang2024large}: Each log message is divided into a static \textbf{Template}, consisting of fixed keywords, and a \textbf{Variable} segment that captures the runtime parameters differing among messages sharing the same template.}
    \label{fig:loghub_flow}
  \end{subfigure}\hfill
  \begin{subfigure}[t]{0.495\linewidth}
    \includegraphics[width=\linewidth]{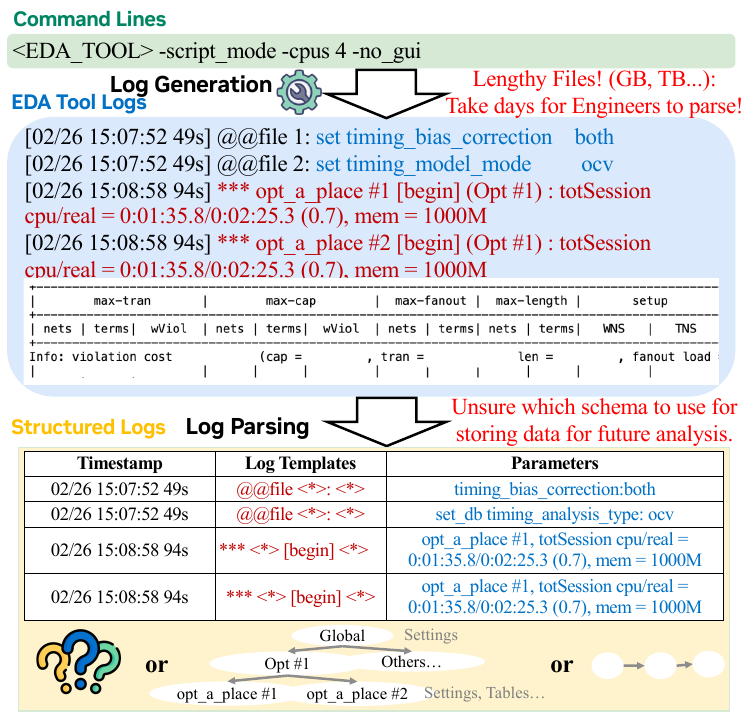}
    \caption{EDA tool logs: Comprising command outputs, progress messages, and embedded performance tables, the logs are partitioned into design \textbf{Stages} (e.g., \texttt{Opt\#1}) and parsed to extract parameter settings along with their corresponding \textbf{Parameters} or \textbf{Tables}.}
    \label{fig:eda_flow}
  \end{subfigure}
  \caption{%
    Two examples of logs schema extraction: (a) LogHub-2.0; (b) EDA Tools Logs.
  }
  \label{fig:combined_flow}
\end{figure*}



Logs record every event in a system—whether high-volume cloud services, large distributed platforms, or embedded devices~\cite{he2021survey,raj2022cloud,khan2016cloud}. By recording every significant action, status update, and error condition, logs empower engineers to diagnose performance bottlenecks, detect anomalies, and reconstruct complex workflows. Yet turning free-form text streams into structured schemas remains difficult. The diversity of log sources and the rapid pace of software evolution continually introduce new formats and conventions, rendering static parsing rules obsolete as soon as they are used.

Electronic Design Automation (EDA) tool logs make things even harder. They mix command-line calls, live progress messages, nested performance tables, and verification proofs all in one stream.
As modern chip designs scale to billions of transistors, span multiple clock domains, and embed deep hierarchies, manually scanning millions of log lines to isolate timing violations, resource usage hotspots, or assertion failures becomes infeasible~\cite{groeneveld2002physical}. Moreover, parsers built on static, hand-crafted patterns, often tuned to benchmarks like LogHub-2.0, struggle to keep pace with evolving tool versions and vendor-specific formats, creating a major bottleneck for scalable, ML-driven log analysis~\cite{he2017drain,du2016spell,wang2009electronic,jansen2003electronic}.

While recent statistical and LLM-based methods have driven notable progress, they remain limited in their ability to handle mixed-content logs. For example, statistical parsers like Drain~\cite{he2017drain} and Spell~\cite{du2016spell} rely on fixed-depth, line-based clustering: they examine each line in isolation, identify recurring tokens, and group messages accordingly. This token-centric view often ignores higher-level semantics and rare but important segments, leading to brittle templates that cannot adapt to evolving formats. Consequently, these approaches struggle with information-sparse or semantically rich logs, such as those produced by EDA tools, where extracting domain-specific parameters and tables requires contextual understanding and background knowledge.

Figure~\ref{fig:combined_flow}(a) illustrates how LogHub-2.0 splits each raw entry into a fixed \emph{Template}, the invariant keywords, and a \emph{Variable} segment capturing runtime parameters. In contrast, Figure~\ref{fig:combined_flow}(b) shows EDA tool logs as a weave of command invocations, progress messages, and nested performance tables partitioned into \emph{Stages} in the design flow. This format often forces engineers into repetitive, time-intensive cycles of manual searching and cross-checking to extract configuration settings and tabular metrics. Moreover, without a predefined schema, even formulating higher-level queries requires first inferring the underlying structure, a process that can itself consume hours or days of effort.

To address these limitations, we introduce \SchemaCoder: the first end-to-end, LLM-driven pipeline for automated log schema extraction that operates without any human-defined regular expressions. 
\SchemaCoder~begins by partitioning log streams into semantically coherent chunks via context-bounded segmentation and uses embedding-based sampling to select representative patterns for analysis. 
A novel hierarchical Question-Tree (Q-Tree) framework orchestrates exploratory, selective, and integrative LLM queries to synthesize robust parsing templates. 
These templates are then refined by a textual-residual-guided evolutionary optimizer.
Furthermore, inspired by the promising performance of gradient boosting technique~\cite{friedman2001greedy, chen2016xgboost} on classical machine learning problems, we developed a novel residual Q-Tree boosting methodology to improve the performance on extracting challenge segments in the log file.
Our contributions are as follows:
\begin{itemize}
    \item To our best knowledge, \SchemaCoder~is the first general, end-to-end, flexible LLM-driven pipeline for log schema extraction that seamlessly handles diverse log types and formats.
    \item We introduce a hierarchical Question-Tree mechanism that flexibly captures critical message patterns and guides the synthesis of parsing templates.
    \item We are the first to formulate log parsing as an optimization problem and employ a residual tree-boosting framework with iterative feedback loops to minimize the loss, thereby enhancing LLM parsing capabilities for rare and challenging log segments.
    \item We perform extensive studies on LogHub-2.0~\cite{jiang2024large}, showing that \SchemaCoder~achieves a 21.3\% average improvement in template accuracy and grouping over state-of-the-art baselines. We further validate its versatility on real-world EDA logs, where its Q-Tree-driven methodology delivers a 57.9\% average boost in pass@k compared to Cursor~\cite{cursor2025}.
\end{itemize}

\section{Related Work}


\begin{figure*}[t]
  \centering
  \includegraphics[width=\linewidth]{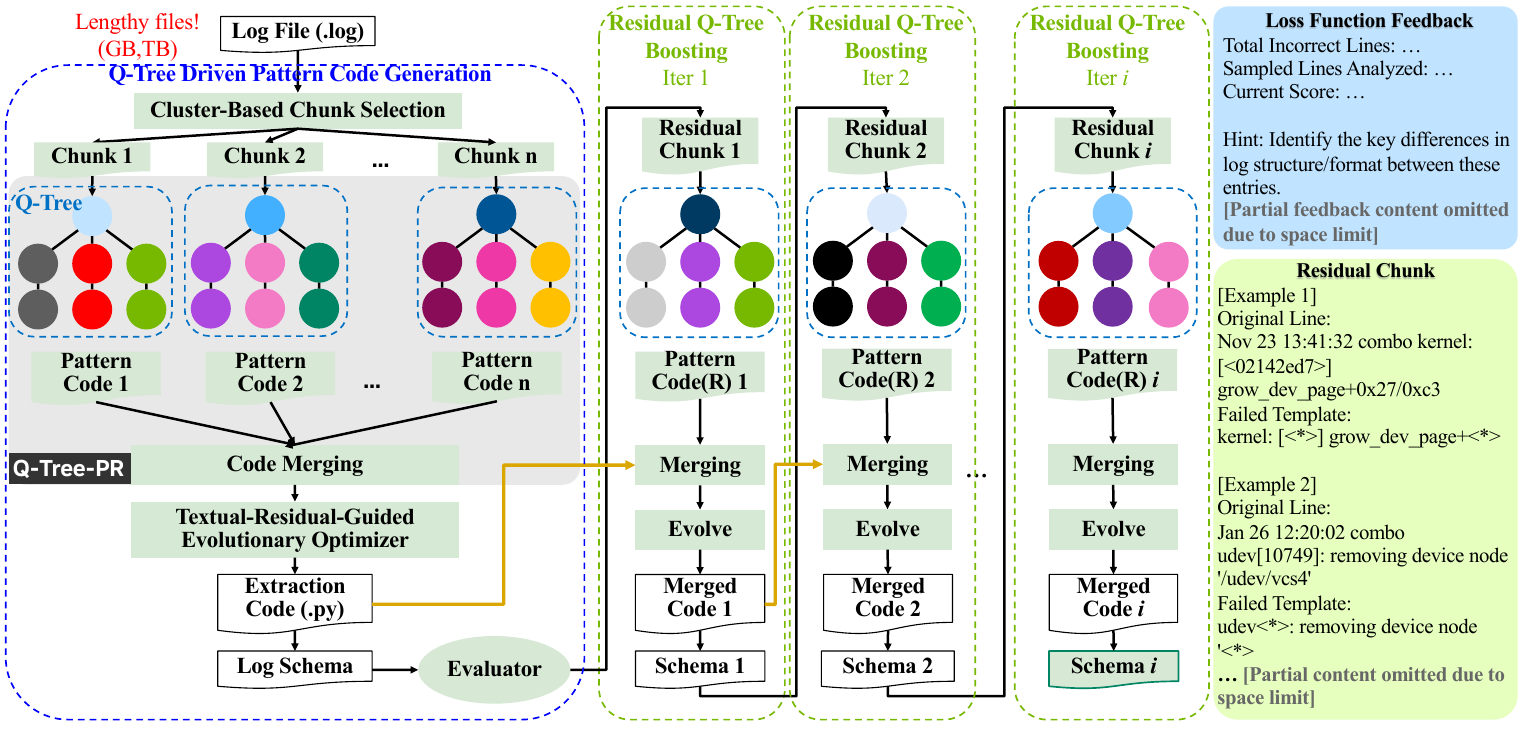}
  \vspace{-0.3cm}
  \caption{The flow of \SchemaCoder:     
  (1) \textbf{Q-Tree Driven Pattern Code Generation}: partitions raw logs into semantic chunks, samples representative chunks, generates pattern code via \QTree, merges these codes, and refines the result with a textual-residual–guided evolutionary optimizer.
  (2) \textbf{Residual Q-Tree Boosting}: iteratively fits pseudo-residual error chunks by sampling challenging chunks and generates additive Q-Tree to correct edge-case failures for robust, high-accuracy schema extraction; 
  In the right side, we show the examples of loss function feedback for textual-residual-guided evolutionary optimizer, and  residual chunk for residual Q-Tree boosting mechanism.}
  \label{fig:schemacoder_pipeline}
\end{figure*}

\subsection{General Log Parsing Techniques}

Traditional log parsers typically follow a two‐step process: normalize timestamps and delimiters, then group messages via token‐based or frequency‐based clustering. Token‐tree methods (e.g., Drain~\cite{he2017drain}) build fixed‐depth parse trees to cluster messages by token position; frequent‐pattern mining approaches (e.g., Spell~\cite{du2016spell}, SLCT~\cite{vaarandi2003data}) identify recurring subsequences or high‐frequency events; graph‐based fitting (e.g., LogMine~\cite{hamooni2016logmine}) models lines and template parameters as a bipartite graph for joint optimization; and robust extensions (e.g., LogCluster~\cite{vaarandi2015logcluster}, SHISO~\cite{mizutani2013incremental}, MoLFI~\cite{messaoudi2018search}) introduce incremental or search‐based heuristics to handle noisy logs. 
Although effective on uniform, well-structured datasets, these pipelines rely on static thresholds and treat each line in isolation. As a result, they become brittle when formats evolve or when handling the interleaved tables and nested progress messages.

LLM‐based approaches treat schema extraction as a generative, context‐aware task. UniParser~\cite{liu2022uniparser} unifies parsing across heterogeneous logs by fusing char‐level embeddings, BiLSTM context encodings, and contrastive learning to label template vs.\ parameter tokens; fine‐tuning strategies (e.g., LogPPT~\cite{le2023log}) combine adaptive random sampling with virtual‐token prompt tuning of RoBERTa to one‐pass tag parameters vs.\ keywords using only a few labeled examples. Although fine‐tuning can achieve high accuracy, it demands substantial computational resources, and prompt‐based variants often lack an end‐to‐end flow for the complex, mixed‐content nature of EDA logs. Some methods are evaluated only on the LogHub‐2k dataset with around 2k lines per log~\cite{zhu2023loghub}, leaving their scalability to larger files untested. \SchemaCoder~addresses this by unifying semantic chunking, multi‐layer Q-Tree queries, automated feedback loops, and residual boosting into a single, scalable pipeline, avoiding expensive re-training or ad hoc heuristics while adapting seamlessly to diverse log formats.

\subsection{LLMs for EDA Applications}
Several recent studies have applied LLMs to EDA tasks, for example, fine-tuned models that generate tool scripts for RTL-to-GDSII flows~\cite{wu2024chateda}, retrieval-augmented prompts that debug Verilog code~\cite{kanagal2025llm}, or vision-guided prompting to parse datasheets into circuit parameters~\cite{chen2024doceda}. These efforts demonstrate that generative models can handle unstructured EDA content, but they each target a single use case and still rely on huge human efforts to customize the pipelines.

To address these challenges, \SchemaCoder~introduces an end-to-end, general framework for handling long-context log files and arbitrary unstructured log formats through a novel Q-Tree-driven pattern code extraction mechanism followed by residual Q-Tree boosting optimization.

\section{Methodology}
We introduce a general, end-to-end, and flexible LLM-driven methodology for log schema extraction. First, we formally define the problem and present the overall workflow of \SchemaCoder. 
Next, we describe the novel Q-Tree-driven pattern code generation and the residual Q-Tree boosting mechanism used to produce the parsing code.
\begin{figure}[t]
  \centering
  \includegraphics[width=0.85\linewidth]{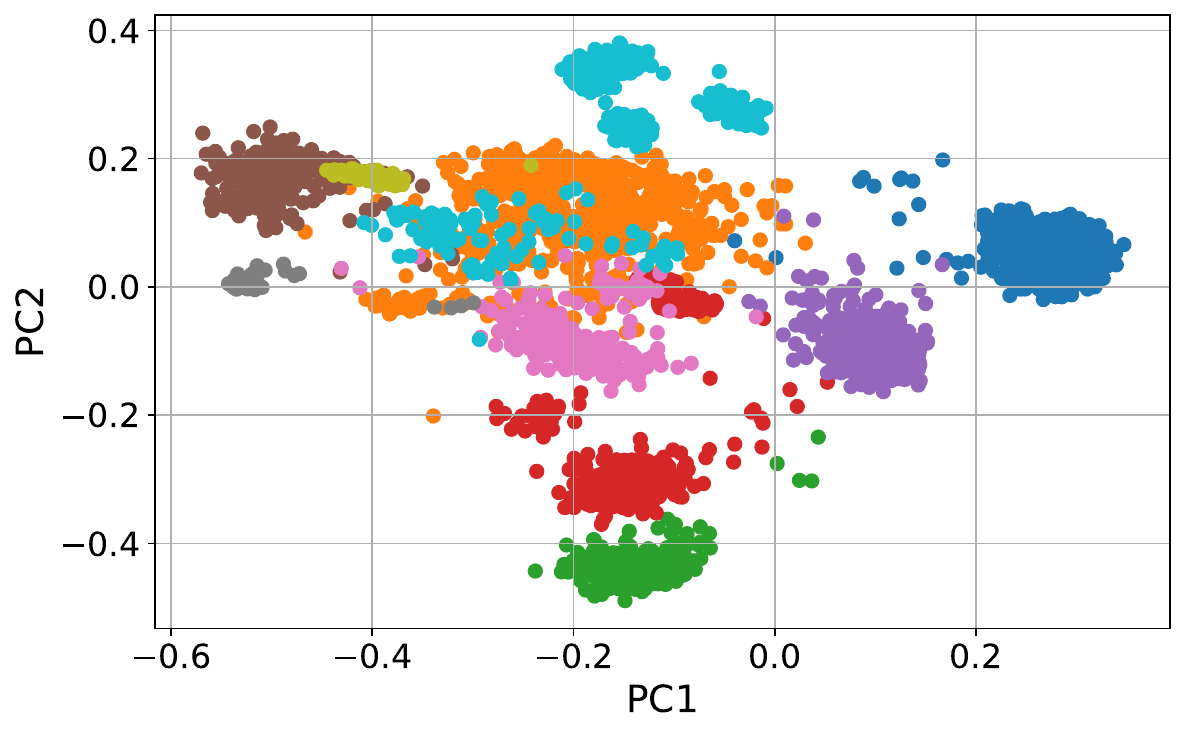}
  \vspace{-0.2cm}
  \caption{PCA projection of log chunk embedding vectors onto the first two principal components (PC1, PC2) on the BGL log. Each color represents a cluster.}
  \label{fig:pca_cluster}
  \vspace{-0.5cm}
\end{figure}


\vspace{-0.2cm}
\subsection{Problem Formulation}
\SchemaCoder~generates code to extract log file schemas from arbitrary unstructured or structured log files. 
We denote \SchemaCoder~as an operator $S(b, m)$, where $b$ represents background knowledge and $m$ is the input log file. 
The output of $S(b, m)$ is the parser code $F$ for log file extraction. 
We define a loss function $L(y, F(m))$ to measure the discrepancy of the extracted log file schema $F(m)$ against the expected schema $y$. 
For loss computation, we adopt the function used in the LogHub-2.0 benchmarks and employ an LLM-based judge when no ground-truth loss is available.
Thus, the problem for each log file $m$ can be formulated as:

\begin{equation}
\hat{F} = \arg \min_{S(b, m)} L\big(y, F(m)\big).
\label{ObjectiveEq}
\end{equation}

\vspace{-0.1cm}
\subsection{Flow Overview}
We propose \SchemaCoder, a unified, end-to-end pipeline for automatic log schema extraction that synthesizes Python parsers via a hierarchical Q-Tree boosting mechanism. 
\SchemaCoder~begins by applying \textbf{cluster-based chunk selection} to partition a raw log file into semantically coherent chunks. 
Each chunk is then fed into our \textbf{Q-Tree-driven pattern code generator}, which generates the initial parser code and then refines the code with the developed textual-residual-guided evolutionary optimizer.
After executing this extractor $F_0$ to obtain an initial log schema, the residual is obtained by $L$ and continue correcting the error residual with the novel~\textbf{residual Q-Tree boosting} mechanism iteratively as depicted in Figure~\ref{fig:schemacoder_pipeline}.
\SchemaCoder~delivers a fully automated log schema extraction pipeline without any human-defined regular expressions.

\vspace{-0.1cm}
\subsection{Cluster-based Chunk Selection Method}

\begin{figure*}[t]
  \centering
  \includegraphics[width=\linewidth]{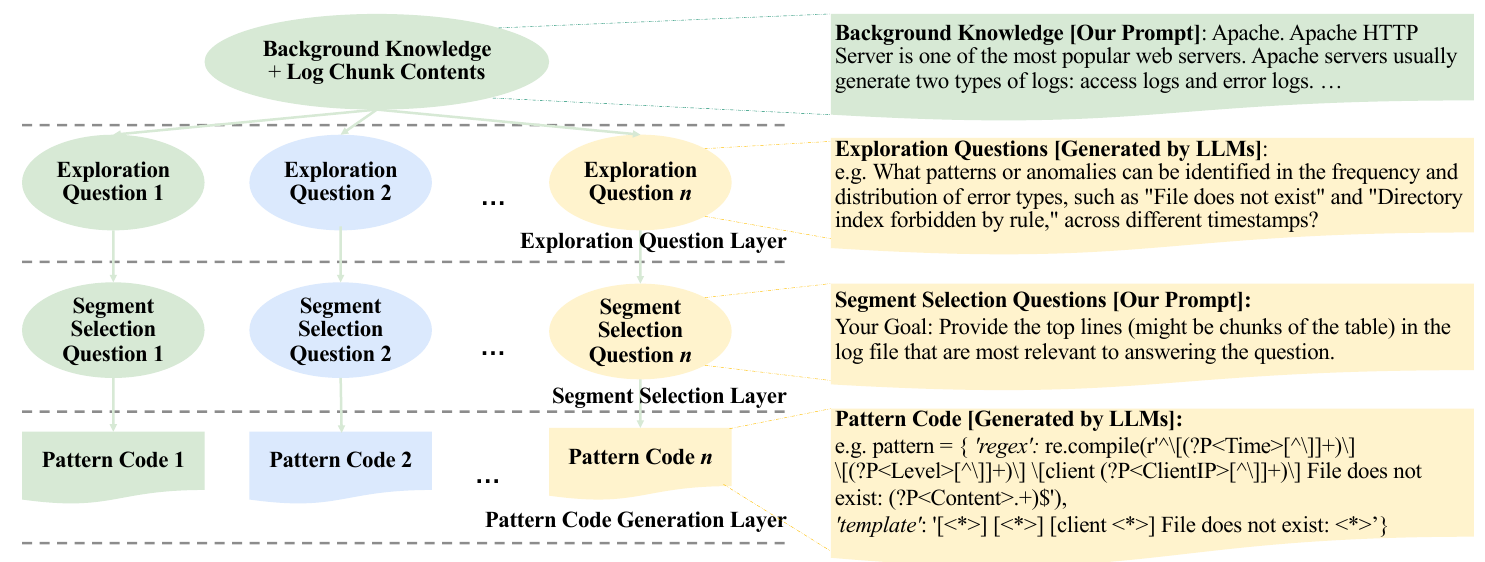}
  \caption{An example hierarchical Q-Tree for Apache log files. The tree structure is fixed to three layers, while each node’s content is generated dynamically by LLMs.}
  \label{fig:qtree}
  \vspace{-0.25cm}
\end{figure*}

First, we divide the log file into chunks suitable for processing within the context window for the following \textbf{Q-Tree-driven pattern code generator}. 
Given the similarity across many chunks in log files, we apply clustering on their embedding representations, and sample three to four representative chunks per cluster for pattern extraction to reduce inference cost while maintaining diversity and coverage.
Figure~\ref{fig:pca_cluster} visualizes the resulting clusters of the BGL log file in the PCA‑reduced embedding space, where colors indicate cluster assignments and markers denote the selected representative chunks.

\begin{figure}[t]
  \centering
  \includegraphics[width=0.85\linewidth]{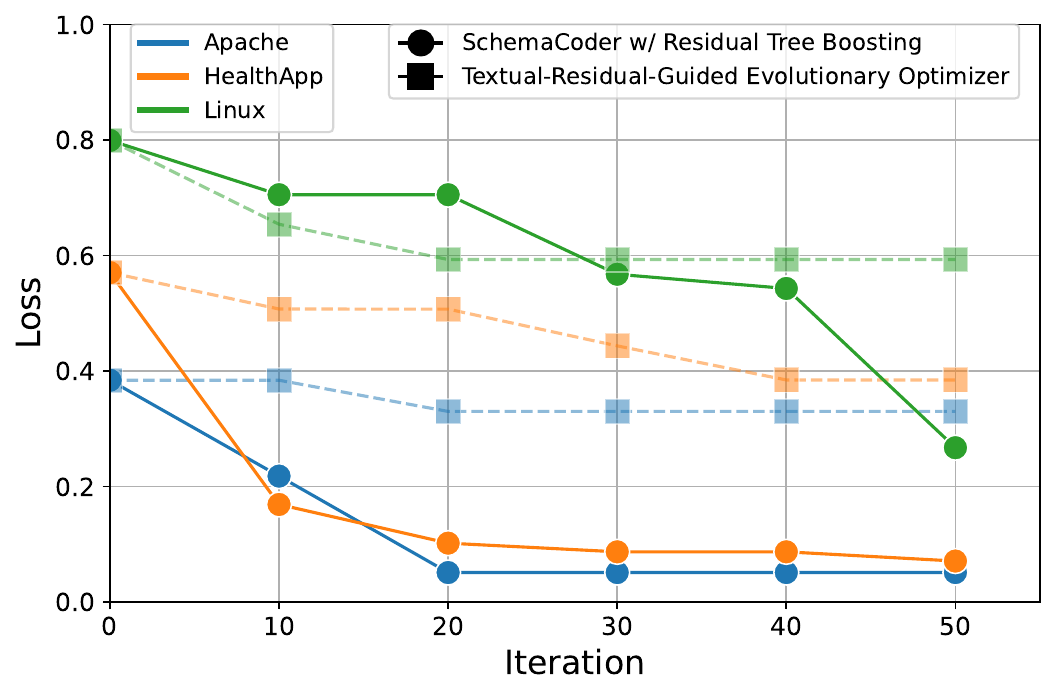}
  \caption{Loss \(L\) over iterations \(i\) on the Apache, HealthApp and Proxifier log, comparing textual-residual-guided evolutionary optimization versus residual Q-Tree boosting in \SchemaCoder. Residual Q-Tree boosting is invoked every 10 iterations, significantly accelerating convergence.}
  \label{fig:residual}
  \vspace{-0.2cm}
\end{figure}


\vspace{-0.1cm}

\subsection{Q-Tree Driven Pattern Code Generation}
We introduce the Q-Tree driven pattern code generation flow with two key components: 1) Hierarchical Q-Tree Pattern Recognition and 2) Textual-Residual-Guided Evolutionary Optimization for generating the parser code.

\noindent \textbf{Hierarchical Q-Tree Pattern Recognition (\QTree)}: 
The hierarchical Q-Tree connects questions directly to pattern code generation.
Given the selected chunks and background knowledge, the proposed Q-Tree generates the pattern code through three fixed layers along each branch as shown in Figure~\ref{fig:qtree}.
Each layer is introduced below.

\begin{itemize}
  \item \textbf{Exploration Question Layer:} \textcolor{black}{We leverage the reasoning capabilities of LLMs to generate open-ended questions from the root node, probing the most salient patterns and anomalies based on the given background knowledge and log chunk contents.}
  \item \textbf{Segment Selection Layer:} 
  We instruct the LLM to pinpoint and extract the most relevant log segments for exploration question while answering it. This targeted “zoom-in” step links initial pattern discovery, making it especially effective for heterogeneous log lines.
  \item \textbf{Pattern Code Generation Layer:} 
  We prompt the LLM to generate template-extraction code for the selected segments. First, it groups lines with similar structures; then, it generates the regular expressions needed to assemble the full pattern-extraction code.
  
\end{itemize}

After generating pattern codes from each question branch, we partition them into groups based on a predefined token limit. 
We then leverage an LLM to iteratively merge these groups, ensuring that all pattern codes are integrated intact into a single parsing program $F$.



\begin{table*}[!t]
  \centering
  \caption{LogHub-2.0 statistics: annotated logs, template counts, and parsing time per application.}
  \label{tab:dataset_stats}
  \resizebox{\textwidth}{!}{%
    \begin{tabular}{lrrrrrrrrrrrrrrr}
      \toprule
      & Apache & BGL & HDFS & HPC & Hadoop & HealthApp & Linux & Mac & OpenSSH & OpenStack & Proxifier & Spark & Thunderbird & Zookeeper & Average \\
      \midrule
      \# Annotated Logs
        & 51,977 & 4,631,261 & 11,167,740 & 429,987 & 179,993 & 212,394 & 23,921 & 100,314 & 638,946 & 207,632 & 21,320 & 16,075,117 & 16,601,745 & 74,273 & 3,601,187 \\
      \# Templates
        & 29 & 320 & 46 & 74 & 236 & 156 & 338 & 626 & 38 & 48 & 11 & 236 & 1,241 & 89 & 249.1000 \\
      Parsing Time (s)
        & 1.5200 & 249.4400 & 394.1700 & 11.1000 & 14.2000 & 7.0100 & 1.2300 & 15.6800 & 14.6400 & 9.2000 & 0.5800 & 59.3300 & 304.1400 & 2.9000 & 77.5100 \\
      \bottomrule
    \end{tabular}%
  }
\end{table*}

\begin{table*}[!t]
  \scriptsize
  \centering
  \setlength{\tabcolsep}{2.8pt}
  \renewcommand{\arraystretch}{1.1} 

  \caption{Average Metrics Comparison on Grouping Accuracy (GA), Parsing Accuracy (PA), F-1 Score of Grouping Accuracy (FGA), and F-1 Score of Template Accuracy (FTA) on different logs across different methodologies}
  \label{tab:ga-transposed}
  \begin{tabular}{l|ccccccccccccccc}
    \toprule
    \textbf{Method}      & Apache & BGL   & HDFS  & HPC   & Hadoop & HealthApp & Linux & Mac   & OpenSSH & OpenStack & Proxifier & Spark & Thunderbird  & Zookeeper & Average \\
    \midrule
AEL & \underline{0.8110} & 0.5183 & 0.7365 & 0.4565 & 0.3833  & 0.2618    & \underline{0.5053} & 0.5100 & 0.5227  & 0.4048    & 0.6838    & \textit{N/A}     & 0.2750      & 0.7728    & 0.5263 \\
    Drain      & \underline{0.8110} & 0.5358 & \underline{0.7910} & 0.4937 & \underline{0.6578}  & 0.2970    & 0.4585 & 0.3540 & \underline{0.6630}  & 0.1975    & 0.4405    & \underline{0.6388} & \underline{0.3388} & 0.8388    & 0.5369 \\
    IPLoM       & 0.5775 & 0.5133 & 0.4205 & 0.4480 & 0.5290  & 0.5683    & 0.4438 & 0.3548 & 0.2878  & 0.1010    & 0.2095    & 0.4460 & 0.4183      & 0.6630    & 0.4272 \\
    LenMa       & 0.5478 & \textit{N/A}      & 0.5928 & 0.3713 & 0.2290  & \textit{N/A}         & 0.4068 & 0.2375 & 0.4725  & 0.3313    & 0.3348    & \textit{N/A}     & \textit{N/A}           & 0.6160    & 0.4140 \\
    LFA         & 0.6425 & 0.3615 & 0.4410 & 0.4962 & 0.5068  & 0.2803    & 0.2690 & 0.3875 & 0.2168  & 0.3380    & 0.1880    & 0.3880 & 0.1520      & 0.5080    & 0.3697 \\
    LogCluster  & 0.1455 & 0.2120 & 0.1395 & 0.3608 & 0.1350  & 0.1905    & 0.2435 & 0.1658 & 0.0645  & 0.1740    & 0.1655    & 0.1038 & 0.1145      & 0.3323    & 0.1819 \\
    LogMine     & 0.6690 & 0.2563 & \textit{N/A}      & \textit{N/A}      & 0.3858  & \textit{N/A}         & 0.4293 & 0.4267 & \textit{N/A}       & \textit{N/A}         & 0.1280    & \textit{N/A}     & \textit{N/A}           & 0.2943    & 0.3699 \\
    Logram      & 0.1745 & \textit{N/A}      & \textit{N/A}      & 0.3275 & 0.0763  & 0.0963    & 0.1190 & 0.1420 & 0.1605  & 0.3353    & 0.0480    & \textit{N/A}     & \textit{N/A}           & 0.5180    & 0.1997 \\
    LogSig      & 0.0475 & \textit{N/A}      & 0.1425 & 0.0823 & 0.0075  & 0.0028    & 0.0365 & 0.0150 & 0.2445  & 0.2838    & 0.2025    & \textit{N/A}     & \textit{N/A}           & 0.1205    & 0.1078 \\
    MoLFI       & 0.3687 & 0.2585 & 0.4755 & 0.3330 & 0.1853  & 0.1515    & 0.1773 & 0.1545 & 0.2825  & 0.0668    & 0.0000    & \textit{N/A}     & \textit{N/A}           & 0.5585    & 0.2510 \\
    SHISO       & 0.4075 & \textit{N/A}      & 0.5593 & 0.0520 & 0.3438  & 0.1440    & 0.1868 & 0.3090 & 0.3290  & 0.3610    & 0.4468    & \textit{N/A}     & \textit{N/A}           & 0.5245    & 0.3331 \\
    SLCT        & 0.2395 & 0.2325 & 0.2108 & 0.3393 & 0.1475  & 0.1710    & 0.0330 & 0.2345 & 0.1728  & 0.5938    & 0.0495    & \textit{N/A}     & \textit{N/A}           & 0.6033    & 0.2523 \\
    Spell       & 0.6055 & \underline{0.5905} & 0.5253 & \textit{N/A}      & 0.1583  & 0.2083    & 0.3870 & 0.2863 & 0.3333  & 0.1955    & 0.2630    & \textit{N/A}     & \textit{N/A}           & 0.6588    & 0.3829 \\
    UniParser   & 0.2730 & \textit{N/A}      & \textbf{0.8645} & 0.6960 & 0.5905  & 0.5250    & 0.2833 & \textbf{0.6488} & 0.2493  & \underline{0.7020}    & 0.3860    & 0.4073 & 0.3363      & 0.7908    & 0.5195 \\
    LogPPT      & 0.6435 & \textit{N/A}      & 0.5710 & \textbf{0.8383} & 0.5725  & \underline{0.8690} & 0.4765 & 0.4200 & 0.3133  & 0.6648    & \underline{0.7925}    & 0.5295 & 0.2910      & \underline{0.8680}    & 0.6038 \\
    \midrule
    \textbf{\SchemaCoder}
                & \textbf{0.9645} & \textbf{0.8365} & 0.7239 & \underline{0.8336} & \textbf{0.8488}  & \textbf{0.9291}    & \textbf{0.7835} & \underline{0.5934} & \textbf{0.7954}  & \textbf{0.8331}    & \textbf{1.0000}    & \textbf{0.8438}     & \textbf{0.4611}           & \textbf{0.9302}    & \textbf{0.8126} \\
    \bottomrule
  \end{tabular}
\end{table*}

\noindent \textbf{Textual-Residual-Guided Evolutionary Optimizer}: We develop an evolutionary algorithm augmented by textual gradient during evaluation phase on top of OpenEvolve framework~\cite{openevolve} for a given initial parser program.  
During evaluation phase, we adopt $\big(1 - L(y, F(m))\big)$ as the score of each generated parser program to provide the gradient of scores for next-generation parser program.
The evolution process integrates MAP-Elites~\cite{mouret2015illuminating}, island-based population models~\cite{romera2024mathematical}, and exploratory program sampling strategies to effectively maintain this balance.
This introspection-augmented EA is novel in log parsing, uniting reflective learning with evolutionary search for prompt optimization.


\subsection{Residual Q-Tree Boosting}
Inspired by the effectiveness and efficiency of gradient boosting tree methods~\cite{chen2016xgboost}, we leverage the additive ensemble of Q-Tree driven pattern code generation to minimize the loss function in Eq.~(\ref{ObjectiveEq}). 
The pseudo-residual for the $i$-th Q-Tree driven pattern code generation iteration is given by:




\begin{equation}
 r_{i} = - \frac{\partial L(y, F_{i-1}(m))}{\partial F_{i-1}(m)} \quad i = 1, \dots, n.
\label{QTreeResidualEq}
\end{equation}

\noindent Here, $r_{i}$ represents the pseudo-residual, corresponding to the residual chunks or lines of the log file $m$ as identified by the loss function $L$. 
We cluster these erroneous chunks or lines and sample $k$ representatives from each cluster for further refinement. 
Next, we generate an additive parser code component $f_i(m)$ using Q-Tree driven pattern code generation to fit the pseudo-residual $r_i$. 
This new component is integrated into the existing parser $F_{i}$ using an LLM-based operator $u(F_i, f_i)$ to produce the updated parser $F_{i+1}$:
\begin{equation}
 F_{i+1}(m) = u\big(F_{i}(m), f_{i}(m)\big).
\label{QTreeEnsembleEq}
\end{equation}
This process is iteratively applied, focusing on the most challenging error patterns, until the loss $L$ reaches zero or a predefined maximum number of boosting iterations is reached.
In Figure~\ref{fig:residual}, both methods start from the same initial parser and track the loss \(L\bigl(y, F_i(m)\bigr)\) at each iteration \(i\). The textual-residual-guided evolutionary optimizer produces a steady but moderate decline in \(L\), whereas residual Q-Tree boosting applies the pseudo-residual correction (Eq.~\ref{QTreeResidualEq}) every ten iterations to generate an additive update (Eq.~\ref{QTreeEnsembleEq}). This targeted residual fitting yields sharper drops in the loss curve, demonstrating faster convergence than evolutionary adaptation alone.

\section{Experiment Results}
We present experimental results applying our framework to LogHub-2.0~\cite{jiang2024large} and EDA logs, comparing its performance against state-of-the-art baselines. First, we conduct extensive studies on LogHub-2.0 and compare the performance of the proposed method with the baselines reported in~\cite{jiang2024large}. Then, we discuss the results of the proposed methodology and Cursor~\cite{cursor2025} on selected QA tasks for EDA log files.
In the following experiments, we use \GPTFo~\cite{openai2024gpt4o} as the core model within \SchemaCoder.

\vspace{-0.1cm}

\subsection{Baselines}
For LogHub-2.0, we compare  \SchemaCoder~against a diverse set of state-of-the-art baselines. The statistical and pattern‑mining methods include AEL~\cite{jiang2008abstracting}, Drain~\cite{he2017drain}, IPLoM~\cite{makanju2009clustering}, Spell~\cite{du2016spell}, SLCT~\cite{vaarandi2003data}, LenMa~\cite{shima2016length}, LFA~\cite{nagappan2010abstracting}, LogCluster~\cite{vaarandi2015logcluster}, MoLFI~\cite{messaoudi2018search}, SHISO~\cite{mizutani2013incremental}, and LogSig~\cite{tang2011logsig}. 
On the other hand, the model or LLM-based methods include LogMine~\cite{hamooni2016logmine}, UniParser~\cite{liu2022uniparser}, and LogPPT~\cite{le2023log}.


Additionally, for the ablation study and EDA Logs, we compare the proposed \QTree~with a leading commercial agentic flow, Cursor~\cite{cursor2025} agent, to write the parser code for LogHub-2.0 benchmarks, and answer the QA tasks given the full EDA log file directly.
Here, we use~\GPTFo~\cite{openai2024gpt4o} as the model in Cursor agent, and default Cursor agent without customizing any cursor rules.

\vspace{-0.1cm}

\subsection{Experiments on LogHub-2.0}
We perform extensive studies of the performance of \SchemaCoder~on log files across distributed system, operating system, supercomputer system, server application, and standalone software.

\vspace{-0.1cm}

\subsubsection{Dataset}
Table~\ref{tab:dataset_stats} reports the total number of annotated log entries and extracted templates for log files in LogHub‑2.0. On average, each system contains 3,601,187 annotated log entries and 249.1 templates.

\vspace{-0.1cm}

\subsubsection{Metrics}
We follow~\cite{zhu2023loghub} and compute Grouping Accuracy (GA), Parsing Accuracy (PA), the F1 score of Grouping Accuracy (FGA), and the F1 score of Template Accuracy (FTA). 
For comparison, we report the average of these four metrics. 
In \SchemaCoder, the loss function \( L(y, F(m)) \) is defined as:  
\begin{equation}
L(y, F(m)) = 1 - \frac{\mathrm{GA} + \mathrm{PA} + \mathrm{FGA} + \mathrm{FTA}}{4}.
\end{equation}


\subsubsection{Experiment Results}
Table~\ref{tab:ga-transposed} shows the average scores of GA, PA, FGA, and FTA across all 14 log files.
Our method achieves an aggregated mean score of 0.8126, outperforming the strongest baseline by over 21.3\%. 
This demonstrates that our approach delivers robust and substantial accuracy improvements across diverse log formats without requiring customization of the parsing flow. 
The average parsing time of the generated parsing code is 77.51 seconds, as shown in Table~\ref{tab:dataset_stats}, while each optimization run of \SchemaCoder~takes approximately four hours on average.

\subsubsection{Ablation Study}
\begin{figure}[t]
  \centering
  \includegraphics[width=0.85\linewidth]{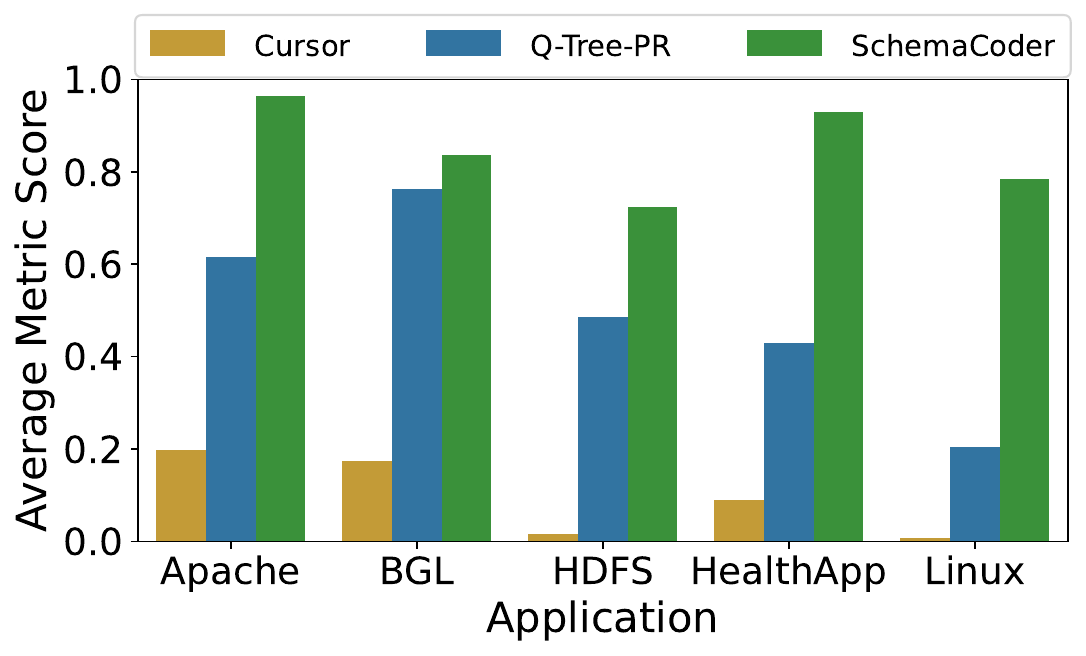}
  \vspace{-0.3cm}
  \caption{Ablation study comparing the log‐processing performance of Cursor, \QTree, and \SchemaCoder~on five representative application logs: Apache, BGL, HDFS, HealthApp, and Linux.}
  \label{fig:ablation}
  \vspace{-0.5cm}
\end{figure}

We study the effectiveness of single \QTree~by comparing the results with Cursor~\cite{cursor2025} and the whole optimization pipeline \SchemaCoder.
We select five representative log files  from distributed systems, supercomputer systems, operating systems, server applications, and standalone software categories.
The single \QTree yields an average 40.19\% improvement in parsing accuracy over the parsing code generated by default Cursor agent on the selected five log files.
With the novel optimization pipeline in \SchemaCoder, the performance is further improved by an additional 34.82\% as shown in Figure~\ref{fig:ablation}. 
Overall, these results demonstrate that while the hierarchical \QTree~lays a strong foundation, the textual-residual-guided evolutionary optimizer and residual boosting is essential for achieving near-perfect parsing accuracy.

\begin{table*}[!h]
\centering
\footnotesize
\begin{tabular}{|p{0.48\linewidth}|p{0.48\linewidth}|}
\hline
\textbf{Logs Description} & \textbf{Question Content} \\
\hline
\multirow{5}{*}{%

\parbox{1\linewidth}{\vspace{1.05pt}This is a physical-design run log capturing the sequential setup, library and floorplan loading, and placement steps—with key warnings, errors, and performance metrics.  

\textbf{Background Knowledge:} Place and route (P\&R) is the back-end process that converts a synthesized netlist into a manufacturable layout.  
Placement assigns exact coordinates to each standard cell and macro, balancing timing, power, and congestion.  
Tools cluster highly connected cells, minimize total wire-length, and legalize positions to prevent overlaps and meet die-outline constraints.}
}
  & Q1. Show me the progression of design density throughout the run. \\ \cline{2-2}
  & Q2. Show me parts of the log where timing slack worsens—i.e., WNS becomes more negative or TNS increases—and pinpoint at which stage timing degradation occurs. \\ \cline{2-2}
  & Q3. Show me all the design options set before running Placement Optimization Stage. \\ \cline{2-2}
  & Q4. Show me all transform committed in the second Design Rule Violation Optimization step and its remaining DRV violations. \\ \cline{2-2}
  & Q5. Show me the progression of WNS, TNS within the first Placement Optimization Stage. \\ \hline
\end{tabular}
\vspace{-0.1cm}
\caption{EDA log descriptions, provided background knowledge and commonly-asked key questions.}
\label{tab:log_questions}
\vspace{-0.2cm}
\end{table*}

    
    

\subsection{Experiment Results on EDA logs}
\begin{figure*}[!ht]
  \centering
  \includegraphics[width=0.83\linewidth]{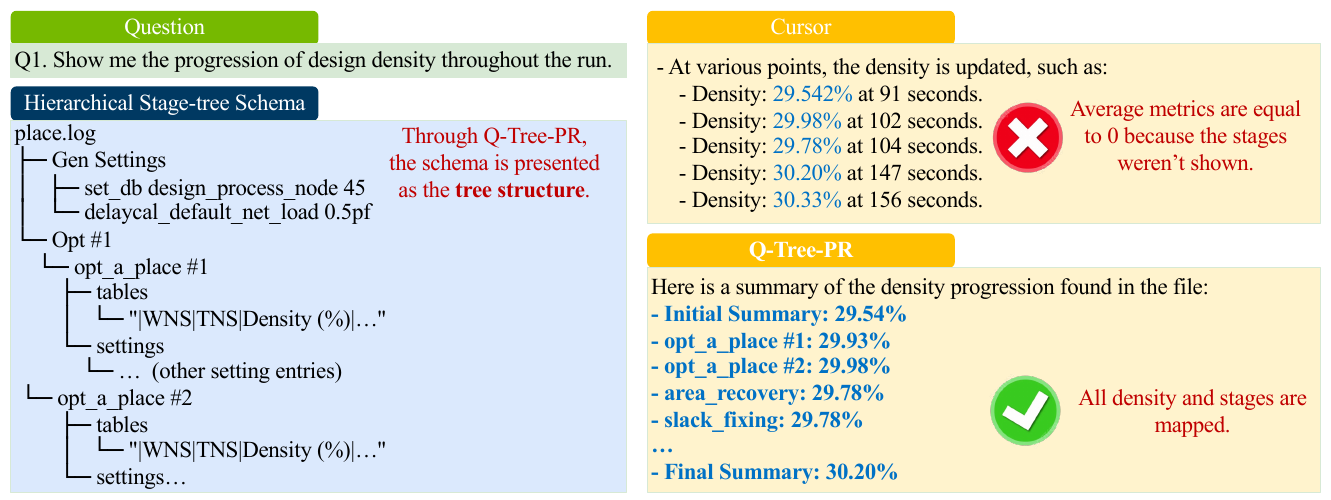}
  \caption{An example of generated answers for Question 1. (Left) The question and extracted hierarchical stage-tree schema from Q-Tree-PR. 
  (Right) The answers generated by Cursor agent and the basic~\GPTFo~agent with the extracted hierarchical stage-tree schema from Q-Tree-PR. Sensitive details omitted to comply with strict vendor licensing requirements.}
  \label{fig:EDA_logs_results}
\vspace{-0.4cm}
\end{figure*}

We study the flexibility and efficiency of the proposed \QTree~procedure for EDA place \& route log files, which include commands, progress messages, and performance tables across stages from floorplanning to placement.
We leverage the \QTree~to extract the key information into a hierarchical stage-tree schema which preserve the execution timeline, and key information (i.e., power-performance-area metrics, settings, etc.) of each optimization stages as shown in~\ref{fig:EDA_logs_results}.
Then, we apply a simple agent (i.e., ~\GPTFo~\cite{openai2024gpt4o}) on top of the extracted log schema for the QA tasks.


\subsubsection{Dataset}


The EDA log file is generated from an industry-leading commercial tool
for Place-and-Route~\cite{cadence2020innovus}.
We carefully select 5 question-answer (QA) tasks from experienced EDA engineers.
The QA tasks covers density progression, slack degradation, settings of different design stages, changes after optimization stages, and stage-specific slack metrics as shown in Table~\ref{tab:log_questions}.

\vspace{-0.2cm}

\subsubsection{Metrics}
For each question, we perform ten independent runs and report pass@1, pass@3, and pass@5 metrics. 
In each run, we compute the hit rate, which is the proportion of correctly retrieved lines, and then average these rates across all trials to obtain the expected pass@k scores. 

\vspace{-0.2cm}

\subsubsection{Experiment Results}


Table~\ref{tab:eda-log-pass135} shows that the \QTree~with basic LLM agent on top can significantly outperforms default Cursor agent on the QA tasks. 
It achieves a pass@1 average of 83.4\% versus the baseline’s 21.1\%, and by pass@5 it reaches near-perfect accuracy compared to the baseline’s 41.6\%. 
This highlights the capability to pinpoint and interpret key commands, metrics, and tables of the proposed method for complex place \& route flow.




\begin{table}[!t]

  \centering
  \caption{Performance Comparison for EDA Logs.}
  \label{tab:eda-log-pass135}
  \resizebox{0.85\columnwidth}{!}{%
    \begin{tabular}{llccc}
      \toprule
      Question & Methodology & pass@1 & pass@3 & pass@5 \\
      \midrule
      \multirow{2}{*}{Q1} & \QTree & \textbf{0.9083} & \textbf{0.9803} & \textbf{0.9956} \\
                         & Cursor & 0.0000 & 0.0000 & 0.0000 \\
      \multirow{2}{*}{Q2} & \QTree & \textbf{0.7500} & \textbf{0.7875} & \textbf{0.7969} \\
                         & Cursor & 0.0000 & 0.0000 & 0.0000 \\
      \multirow{2}{*}{Q3} & \QTree & \textbf{0.9578} & \textbf{0.9925} & \textbf{0.9987} \\
                         & Cursor & 0.4244 & 0.6749 & 0.8049 \\
      \multirow{2}{*}{Q4} & \QTree & \textbf{0.7000} & \textbf{0.9250} & \textbf{0.9813} \\
                         & Cursor & 0.2667 & 0.4713 & 0.5369 \\
      \multirow{2}{*}{Q5} & \QTree & \textbf{0.8526} & \textbf{0.9914} & \textbf{0.9991} \\
                         & Cursor & 0.3643 & 0.6524 & 0.7400 \\
      \bottomrule
    \end{tabular}%
  }
\end{table}


Figure~\ref{fig:EDA_logs_results} shows a representative run for Question 1.
On the left, \QTree~extracts the key information into a hierarchical stage-tree schema that captures key stages (e.g., \texttt{opt\_a\_place \#1}) in execution order along with their corresponding density tables. 
On the right, we show the answers of Cursor and \QTree~for this task.
Cursor~\cite{cursor2025} fails to identify any design stages and yields zeroed metrics. 
In contrast, with the extracted hierarchical stage-tree schema, the basic~\GPTFo~agent can correctly locate each design stage and report the density values in order. 

\section{Conclusion}



In summary, we introduce \SchemaCoder—the first end-to-end, LLM-driven log schema extractor that requires no manual formatting. By uniting context-bounded segmentation, embedding-based sampling, hierarchical Question-Tree synthesis, and residual boosting, \SchemaCoder delivers a 21.3\% lift on LogHub-2.0 and a 57.9\% pass@k increase on challenging EDA logs. Going forward, we will develop domain-specific feedback loops to ensure robust, reliable evaluation on real-world industrial logs.
Future work includes extending the framework to real-world, dynamic, and heterogeneous industrial logs, such as those generated from hardware design, manufacturing, etc.

\section*{Acknowledgement}
We thank our collaborators Yong Liu, Xiaofen Xu from Cadence for providing valuable real-world use cases.

\bibliography{aaai2026}

\newpage

\clearpage
\onecolumn


\section*{Supplementary material}

# Appendix: SCHEMACODER: Automatic Log Schema Extraction Coder with Residual Q-Tree Boosting

## Technical Appendix

In this Technical Appendix, we present the essential implementation details of the SCHEMACODER framework. To uphold the highest standards of reproducibility and transparency, we will publicly release all source code and datasets associated with our experiments, empowering researchers to effortlessly reproduce and rigorously validate every result reported in this paper.

## Additional Details about Experimental Setup

**Cluster-based Chunk Selection**    We set the maximum cluster count to $k_{\max} = 11$ for typical logs and to $k_{\max} = 41$ for exceptionally large logs (e.g., BGL, Thunderbird, Spark). Each log is first divided into contiguous 1,000-line segments—truncated to 800 lines for very long entries—so that every chunk fits within the GPT-4O context window (OpenAI 2024). We embed each segment using the pretrained `all-MiniLM-L6-v2` model (Reimers and Gurevych 2019) and then apply k-means clustering, sweeping $k$ from 2 up to $k_{\max}$ and selecting the value that maximizes the average silhouette score. Segments are assigned to their respective clusters, and we randomly sample two representative chunks from each cluster for downstream analysis. Because EDA logs are heterogeneous, we will use all the chunks to complete the following steps.

**Hierarchical Q-Tree Pattern Recognition (Q-TREE-PR)**    We use GPT-4O with the full version of `gpt-4o-2024-11-20` (OpenAI 2024) . The temperature we use is 0.7 to enhance the answer's variability. All the prompts are presented in the following section Prompt Details. We collected the statistics for the tokens and the time used for Q-TREE-PR and code merging as Table 1 shows. Table 1 shows that our Q-TREE-PR pipeline slashes the number of chunks sent to GPT-4O by two to three orders of magnitude—transforming logs with thousands of original segments into just a few dozen representative examples—thereby reducing token usage and total runtime proportionally. Here, the *Compression Ratio* is calculated as the number of representative chunks divided by the number of original chunks ($\frac{\#\text{Chunks}}{\#\text{Original Chunks}}$). Smaller logs complete in just a few seconds, while even multi-million-token workloads finish within minutes, and this efficiency holds across both homogeneous system logs. By collapsing redundant patterns before generation and maintaining a roughly linear relationship between token volume and processing time, Q-TREE-PR delivers fast, scalable parsing without sacrificing the critical information needed for downstream schema extraction.

**Residual Tree Boosting**    For our LogHub-2.0 experiments, we configure the inner loop of the Textual-Residual-Guided Evolutionary Optimizer to run for 10 iterations, while the outer residual Q-Tree boosting loop is set to 20 iterations to allow sufficient exploration of residual patterns without excessive computation. We terminate the process once the improvement over three consecutive iterations falls below 0.1%, a threshold chosen to balance diminishing returns against runtime overhead.

In our evolutionary algorithm (Sharma 2025), we leverage two Claude variants—Claude-3.5 (version *claude-3-5-sonnet-20240620*) and Claude-3.7 (version *claude-3-7-sonnet-20250219*). At each generation, we sample from Claude-3.5 with probability 0.7 and from Claude-3.7 with probability 0.3, striking a balance between the stability of the former and the enhanced capabilities of the latter.

## Prompt Details

In this subsection, we detail the prompt templates that drive our Q-TREE-PR, each carefully made to enforce structure, clarity, and domain relevance. By coupling system and user messages with explicit background context, task objectives, constraints, and output formats, we ensure that every stage—from question formulation to final program assembly—operates with reproducible rigor and maximal effectiveness.

Table 1: Chunk Counts, Tokens, and Time Used for Q-Tree-PR by Application

| Application | # Original Chunks | # Clusters | # Chunks | Compression Ratio | # Tokens | Time (s) |
|---|---|---|---|---|---|---|
| Apache | 52 | 2 | 4 | 0.0770 | 505 321 | 10.4515 |
| Proxifier | 22 | 6 | 12 | 0.5450 | 1 334 407 | 26.3205 |
| HDFS | 11 168 | 10 | 20 | 0.0020 | 1 524 874 | 24.7794 |
| Hadoop | 225 | 4 | 8 | 0.0360 | 1 324 992 | 925.6134 |
| Linux | 24 | 9 | 18 | 0.7500 | 505 321 | 10.7663 |
| Mac | 101 | 8 | 16 | 0.1580 | 2 726 733 | 2 268.4922 |
| OpenSSH | 639 | 4 | 8 | 0.0130 | 1 069 502 | 20.2550 |
| OpenStack | 1 039 | 10 | 20 | 0.0190 | 3 012 456 | 562.7841 |
| Spark | 16 076 | 20 | 40 | 0.0025 | 3 953 120 | 83.2574 |
| Thunderbird | 16 602 | 13 | 26 | 0.0020 | 2 064 321 | 52.1386 |
| Zookeeper | 75 | 3 | 6 | 0.0800 | 1 061 331 | 965.8810 |
| HealthApp | 213 | 2 | 4 | 0.0190 | 491 645 | 543.4053 |
| HPC | 430 | 8 | 16 | 0.0370 | 1 418 559 | 36.9856 |
| BGL | 5 790 | 11 | 22 | 0.0040 | 1 725 432 | 45.6273 |

**Exploratory Question Layer (Figure 1)**   This template tasks the model with producing a fixed number of domain-relevant "What" or "How" questions to uncover key log features. By embedding background knowledge and strict relevance constraints, each question extends beyond the input and targets a distinct analytical dimension. The enforced output format (`Question 1: ...`) and the requirement for domain-specific terminology sharpen focus and lay the foundation for downstream selection.

**Segment Selection Layer (Figure 2)**   Building on the exploratory questions, this template instructs the model to extract the most relevant log lines or contiguous chunks. It balances precise selection criteria—favoring concrete metrics, structured sections, and contextual completeness—with an upper limit on excerpts to avoid overload. The standardized output (`Pattern 1: ...`) distills raw logs into a concise evidence set for the next stage.

**Pattern Code Generation Layer (Figure 3)**   At this stage, the model transforms selected snippets into valid Python code that extracts the defined schema patterns. The prompt breaks the task into clear steps—identifying fields, implementing extraction logic, and marking `EVOLVE-BLOCK` regions for iterative refinement—ensuring syntactic correctness, generality, and maintainability. This yields an immediately executable parser scaffolded for future optimization.

**Code Merging Step (Figure 4)**   Finally, this template merges independently generated parsing modules into a unified, production-ready program. It focuses on consolidating shared logic, removing redundancies, and integrating comprehensive error handling to cover all anticipated log patterns. Wrapping the core pipeline in `EVOLVE-BLOCK` markers preserves flexibility for ongoing improvements, while enforcing clean, efficient Python syntax guarantees robustness and ease of integration.

Here are some explanations and examples for the variables in the prompts: For the variable parts in the prompt 1, the detailed information is as follows:

- **`<Background Knowledge>`**: It is the background knowledge for the log file with an example of the input and output of the targeted pattern code. We collect it from the paper (Zhu et al. 2023).. For example, the Apache log's background knowledge is: Apache. Apache HTTP Server is one of the most popular web servers. Apache servers usually generate two types of logs: access logs and error logs. This dataset provides an error log for the purpose of research on anomaly detection and diagnosis. The log file was collected from a Linux system running Apache Web server, as part of the Public Security Log Sharing Site project. The input and output example is as follows:

```
1    Logs Chunk to be parsed (It is not the code input. It is a chunk from the log path.
         The code input is nothing):
2    [Thu Jun 09 06:07:19 2005] [notice] jk2\_init() Found child 2336 in scoreboard slot
         6
3    Output (When calling, it should be structured\_data, templates = parse\_log()):
4    structured\_data (list of dicts):
5    [{
6        \'LineId\': 1,                          # Line number (1-based)
7        'Time': 'Thu Jun 09 06:07:19 2005',     # Combined date + time
8        'Level': 'notice',                      # Log level
9        'Content': 'jk2_init() Found child 2336 in scoreboard slot 6',  # Original log
             content
10       'EventId': 'a1b2c3d4',                  # 8-char hash of template
```

```
11              'EventTemplate': 'jk2_init() Found child <*> in scoreboard slot <*>'  #
                  Template (variables      <*>)
12          },...]
13      templates (dict):
14      {
15          "a1b2c3d4": 1,
16      }
```

- **`<num exploratory>`**: The number of branches is predefined. In our experiments, the number for LogHub-2.0 is 2 while the EDA logs is 10 because it contains more information.

- **`<log chunk content>`**: This refers to the selected log chunk after segmentation.

For the newly added variable parts in the prompt 2, the detailed information is as follows:

- **`<parent node question>`**: It refers to the exploratory question generated from the parent node from the Exploratory Question Level.

For the newly added variable parts in the prompt 3, the detailed information is as follows:

- **`<schema pattern>`**: This refers to the schema patterns from the answer of 'Prompt Code Generation' Layer.

- **`<code output format>`**: This refers to a unified pattern for evaluation. The format is like:
Code Output Format:

```
 1  # EVOLVE-BLOCK-START
 2  def parse_log():
 3      # Define the log path
 4      # Read the log file
 5      # RECOGNIZED-BLOCK-START
 6      # Define general extraction patterns for each log format from the log path
 7      # RECOGNIZED-BLOCK-END
 8      # Output the structured data to a csv file
 9      return structured_data
10  # EVOLVE-BLOCK-END
```

For the newly added variable parts in the prompt 4, the detailed information is as follows:

- **`<all codes>`**: This refers to all codes as a list from all the branches of the Pattern Code Generation Layer.

For the newly added variable parts in the prompt 5, the detailed information is as follows:

- **`<log lines number>`**: This refers to the line number of the log file.

- **`<templates number>`**: This refers to the template number extracted from the current pattern code parsing on the log file.

- **`<Textual Residual Feedback>`**: This refers to the feedback given the evaluator, including some specific failed cases when doing the log parsing. Also the Residual Chunk part will be saved as the input of the Residual Question Tree as well.


## LogHub-2.0 Examples

**Q-Tree Output Examples**     In Figure 7, we showcase two distinct Q-Tree branches applied to an Apache log chunk enriched with background knowledge: the left branch formulates an exploration question targeting "Directory index forbidden by rule" messages, automatically selects representative log lines, and generates a regex extractor that captures timestamp, client address, and error descriptor; the right branch focuses on "File does not exist" events, isolates the relevant entries with file paths, and synthesizes a pattern-matching routine to extract time, client, and missing path. These examples demonstrate how our pipeline adapts to different line patterns by combining LLM-driven query formulation with precise regex-based code generation for robust, scalable log parsing.


## EDA Log Examples

**Q-Tree Output Examples**     In Figure 8, we illustrate our Q-Tree pipeline applied to a single, background-enriched EDA log chunk: the process begins with a domain-aware exploration question (e.g., "What is the structure of metric tables and their column semantics?"), proceeds to automatic selection of the most relevant log spans (such as stage/sub-stage markers and tabular snapshots), and culminates in the synthesis of concise regular-expression routines that capture the underlying schema. By combining LLM-driven schema inference with precise pattern matching, this layered approach delivers robust, scalable extraction across the heterogeneous formats typical of EDA tool logs.

**System Message**: You are an expert category engineer applying background knowledge to synthesis log analysis. Always format questions clearly and ensure they start with 'What' or 'How'. Generate exactly **<num_exploratory>** diverse questions that explore different aspects of the log content.

**User Message**:
—
** Background Knowledge: **<Background Knowledge>**
—
Context: You have been provided with:
* The log file
* Background knowledge on the log
Your Goal:
- Formulate **<num_exploratory>** important exploratory questions that should be asked to be able to identify key features and then extract the pattern to match the key features.
Your question might:
- Build directly on the theoretical concepts from the background knowledge.
- Focus on what distinguishes this log from other quality levels.
- Each question should explore a different aspect of the log content.
- Generate diverse questions that cover different log analysis dimensions.
Constraints:
- Ensure the question is directly relevant to the log file.
- Ensure the question is directly derived from the background knowledge.
- If necessary, use domain-specific terminology and aim to retrieve domain-specific knowledge.
- Avoid simply restating the information already provided or present in the background knowledge; instead, aim to advance the user's understanding or resolution of the problem.
- Each question should be unique and explore different aspects of the log content.
- Your response should be of the following format:'Question 1: Q' (without quotes) where Q is your proposed exploratory question.
**<num_exploratory>** questions, each exploring a different dimension of the log analysis.
- Do not write anything other than the question format specified above.
—
Here is the log content:
**<log_chunk_content>**
—
Output Format:
- Question 1: Q1
- Question 2: Q2
- Question 3: Q3
...
- Question **<num_exploratory>**: Q**<num_exploratory>**

Figure 1: Prompt template for the Exploration Question Layer of Q-TREE-PR.

**Examples on EDA Log Contents**    In Figure 9 we see a snippet of an EDA tool log that mixes simple text lines, small tables of placement and timing numbers, script calls and checksum entries. Because these logs jump between different layouts—headers, pipes-delimited tables, file paths and multi-line messages—traditional rule-based parsers often fail. Our ML parser instead learns to tag five core phases: startup and area reporting, placement iterations (showing quality metrics), script/checksum checkpoints, timing-optimization messages, and resize/buffer tables. Once tagged, downstream ML models can easily pull out trends (e.g., how placement quality improves over iterations), flag unusual values (e.g., sudden overflow spikes), point errors to the exact phase for quick debugging, and gather resource stats (e.g., buffer counts) to predict design costs early on.

**System Message**: You are an expert logs engineer. Generate segments selection in the specified output format. Focus on identifying log structure and data patterns.

**User Message**:

—

** Background Knowledge: **\<Background Knowledge\>**

—

** Problem: **\<parent_node_question\>**

—

Context: You have been provided with:

* The log file
* A problem related to that log file

Your Goal:

- Provide the top lines/chunks in the log file that are most relevant to answering the question.

Guidelines:

- Focus on lines/chunks that contain specific metrics, measurements, or data points related to the question.
- Look for recurring patterns, table headers, or structured data sections that indicate important log events.
- Consider both individual lines and contiguous chunks/sections that work together to provide complete information.
- Think about what makes these lines/chunks distinctive and essential for answering the parent question.
- Consider the context around selected lines/chunks to ensure they provide complete information.

Constraints:

- Generate one pattern per question.
- Select no more than 5-10 of the most critical lines/chunks or line chunks to avoid information overload.
- Ensure selected lines/chunks actually contain concrete, answerable information rather than generic text.
- Avoid selecting redundant or duplicate information unless it shows important patterns.
- Prioritize lines/chunks that directly address the core question rather than tangential information.
- For the line_content or chunk_content part, provide the full content of the line, not just a part of it.

Output Format:

Provide your response as a string object with this exact structure:

Pattern 1:

- \<section_1_1_content\>
- \<section_1_2_content\>
- \<section_1_3_content\>
- ...

Pattern 2:

- \<section_2_1_content\>
- \<section_2_2_content\>
- \<section_2_3_content\>
- ...

—

Here is the log content: **\<log_content\>**

Figure 2: Prompt template for the Segment Selection Layer of Q-TREE-PR.

**System Message:** You are an expert log engineer. You are given a list of target log chunks to extract the useful information from the target log chunks to answer the question. You need to generate code that will extract the information from the target log chunks according to the schema pattern. Always wrap your code with EVOLVE-BLOCK-START and EVOLVE-BLOCK-END markers.

**User Message:** You need to generate code that will extract the information from the target log chunks according to the schema pattern.

Background Knowledge: **<Background Knowledge>**

Task:

1. Identify the useful information from the target log chunks to answer the question.

2. Generate code that will extract the information from the target log chunks according to the schema pattern.

3. The code should be valid Python syntax that can be directly implemented.

4. The code should be able to handle the target log chunks and extract the information from the target log chunks according to the schema pattern.

5. Please wrap any code intended for future refinement within EVOLVE-BLOCK markers to ensure seamless integration. Use EVOLVE-BLOCK-START and EVOLVE-BLOCK-END to delineate the sections you plan to evolve or improve. If you're uncertain about the exact boundaries, feel free to enclose a larger block of code within these markers.

Rules:

1. The code should use general patterns to extract the information from the log.

2. Input should be the log chunks

3. Please start the line number from 1

Log Chunks And Schema Patterns:

**<schema_pattern>**

Please generate the code in the following format:

Code Output Format:

**<code_output_format>**

Instruction: Generate the initial program.

Figure 3: Prompt template for the Pattern Code Generation Layer of Q-Tree-PR.

**System Messages**: You are an expert log parsing engineer specialized in merging and unifying code segments. Generate clean, efficient, and comprehensive parsing code that combines all provided segments while eliminating redundancies.

**User Messages**: You are an expert log parsing engineer. You need to merge multiple code segments that were generated from different chunks of the log file into a single, unified, and comprehensive parsing program.

Background Knowledge:

**<Background Knowledge>**

Task:

1. Analyze all the provided code segments below.

2. Identify common patterns and complementary functionalities.

3. Merge them into a single, unified parsing program.

4. Eliminate redundancies while preserving all unique parsing capabilities.

5. Ensure the final code is well-structured, efficient, and maintainable.

6. Include comprehensive error handling and logging.

7. Make sure the merged program can handle all the different log patterns found across chunks.

Requirements:

1. The merged code should be valid Python syntax that can be directly executed

2. Use EVOLVE-BLOCK-START and EVOLVE-BLOCK-END markers around the main parsing logic

3. Return a complete, runnable Python program

Individual Code Segments To Merge:

**<all_codes>**

Code Output Format:

**<code_output_format>**

Instruction: Generate the merged program.

Figure 4: Prompt template for the Code Merging step.

Evaluated **`<log lines number>`**
Generated **`<templates number>`** templates


**Residual Chunk:**
**`<Textual Residual Feedback>`**

Figure 5: Loss Function Feedback and Residual Chunk.

---

Found 51978 wrong lines, sampled 20 for analysis:
Lines with Wrong Templates:
1. Original Line: N/A — Content: [Sun Dec 04 05:15:09 2005] [error] [client 222.166.160.184] Directory index forbidden by rule: /var/www/html/ — Failed Template: [* ] [* ] [client * ] Directory index forbidden by rule: *
2. Original Line: N/A — Content: [Mon Jan 09 17:36:57 2006] [error] [client 218.154.80.8] script not found or unable to stat: /var/www/cgi-bin/ip.cgi — Failed Template: [* ] [* ] [* ] script not found or unable to stat: *
3. Original Line: N/A — Content: [Thu Nov 03 16:17:41 2005] [error] [client 212.45.53.176] File does not exist: /var/www/html/b2evo — Failed Template: [* ] [* ] [client * ] File does not exist: *
4. Original Line: N/A — Content: [Thu Sep 29 08:51:51 2005] [error] [client 61.146.41.126] Directory index forbidden by rule: /var/www/html/ — Failed Template: [* ] [* ] [client * ] Directory index forbidden by rule: *
5. Original Line: N/A — Content: [Fri Feb 10 17:40:39 2006] [error] [client 144.202.242.108] File does not exist: /var/www/html/blogs — Failed Template: [* ] [* ] [client * ] File does not exist: *
6. Original Line: N/A — Content: [Sun Nov 20 05:16:17 2005] [notice] jk2_init() Found child 21338 in scoreboard slot 0 — Failed Template: [* ] [* ] jk2_init() Found child * in scoreboard slot *
7. Original Line: N/A — Content: [Thu Sep 29 15:18:07 2005] [error] [client 172.152.131.48] File does not exist: /var/www/html/scripts/root.exe — Failed Template: [* ] [* ] [client * ] File does not exist: *
8. Original Line: N/A — Content: [Tue Jan 24 00:41:47 2006] [error] [client 209.250.116.251] File does not exist: /var/www/html/mambo — Failed Template: [* ] [* ] [client * ] File does not exist: *
9. Original Line: N/A — Content: [Tue Jan 31 23:40:05 2006] [error] [client 65.110.43.170] File does not exist: /var/www/html/blogs — Failed Template: [* ] [* ] [client * ] File does not exist: *
10. Original Line: N/A — Content: [Tue Jan 10 05:02:55 2006] [error] [client 163.117.142.142] File does not exist: /var/www/html/blog — Failed Template: [* ] [* ] [client * ] File does not exist: *
11. Original Line: N/A — Content: [Mon Nov 21 23:41:05 2005] [error] mod_jk child workerEnv in error state 4 — Failed Template: [* ] [* ] mod_jk child workerEnv in error state *
12. Original Line: N/A — Content: [Mon Nov 14 00:17:47 2005] [error] [client 213.203.171.98] File does not exist: /var/www/html/blog — Failed Template: [* ] [* ] [client * ] File does not exist: *
13. Original Line: N/A — Content: [Sun Dec 04 07:02:01 2005] [notice] workerEnv.init() ok /etc/httpd/conf/workers2.properties — Failed Template: [* ] [* ] workerEnv.init() ok *
14. Original Line: N/A — Content: [Thu Nov 10 10:17:16 2005] [error] [client 64.34.162.99] script not found or unable to stat: /var/www/cgi-bin/awstats.pl — Failed Template: [* ] [* ] [* ] script not found or unable to stat: *
15. Original Line: N/A — Content: [Mon Jul 11 23:32:38 2005] [error] [client 61.109.224.194] File does not exist: /var/www/html/_vti_bin — Failed Template: [* ] [* ] [client * ] File does not exist: *
16. Original Line: N/A — Content: [Sat Nov 19 03:30:36 2005] [error] [client 200.24.17.195] File does not exist: /var/www/html/xmlsrv — Failed Template: [* ] [* ] [client * ] File does not exist: *
17. Original Line: N/A — Content: [Thu Feb 16 01:00:28 2006] [error] mod_jk child init 1 -2 — Failed Template: [* ] [* ] mod_jk child init * *
18. Original Line: N/A — Content: [Wed Nov 09 01:54:22 2005] [error] [client 203.218.126.117] Directory index forbidden by rule: /var/www/html/ — Failed Template: [* ] [* ] [client * ] Directory index forbidden by rule: *
19. Original Line: N/A — Content: [Sun Feb 26 21:00:24 2006] [error] [client 194.90.30.83] File does not exist: /var/www/html/mambo — Failed Template: [* ] [* ] [client * ] File does not exist: *
20. Original Line: N/A — Content: [Tue Jan 10 04:33:43 2006] [error] [client 202.124.131.59] File does not exist: /var/www/html/sumthin — Failed Template: [* ] [* ] [client * ] File does not exist: *

Figure 6: An example of **`<Textual Residual Feedback>`**.

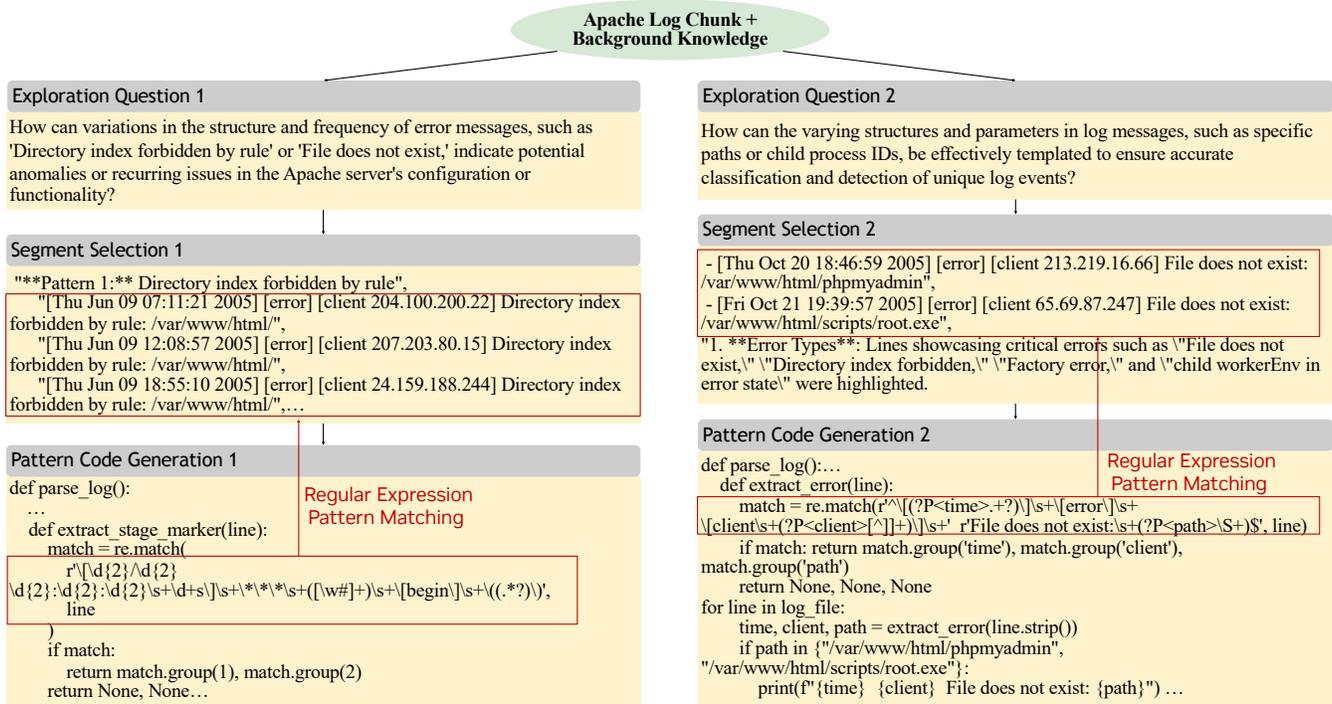

Figure 7: Apache Log Q-Tree Examples. The pattern code is partially shown with the regular expressions to match the segments.

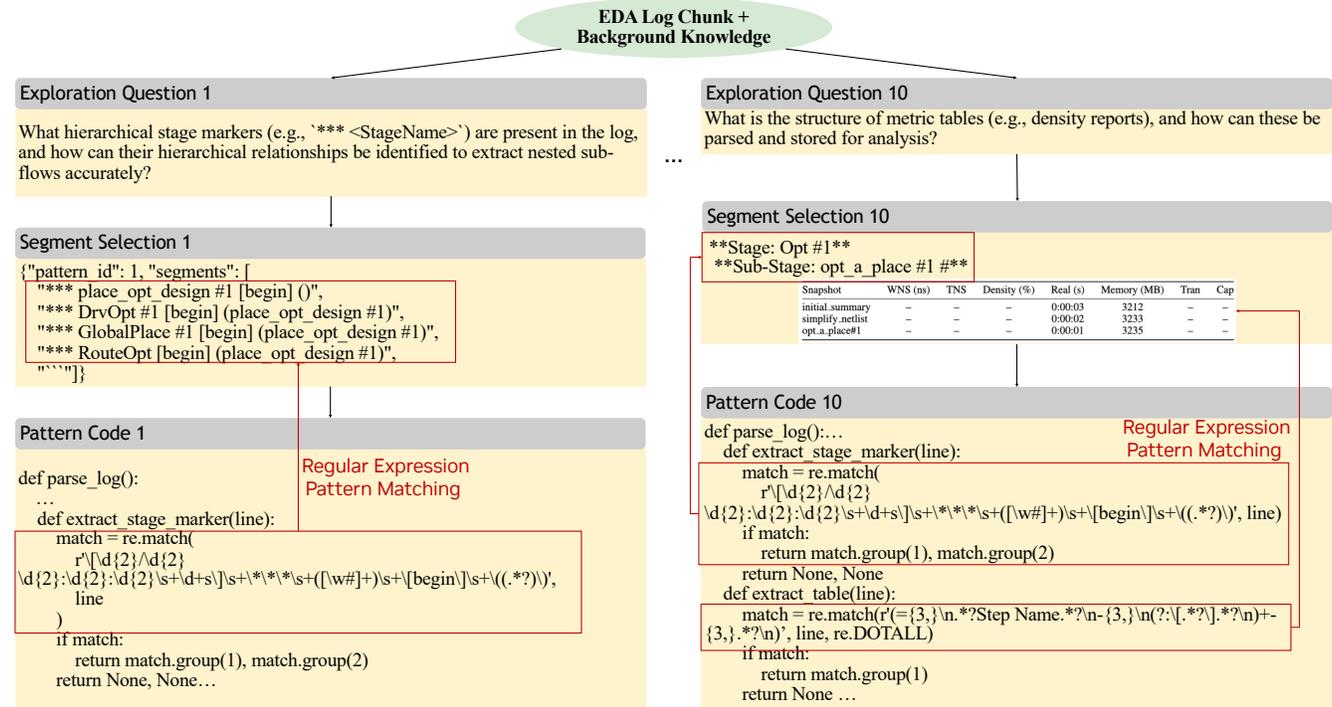

Figure 8: An EDA Log Q-TREE-PRExamples . The pattern code is partially shown with the regular expressions to match the segments.

```
1  [2025-07-16T19:42:51.950Z]
   ======================================================================
2  [2025-07-16T19:42:51.950Z] floorplan final report_design_area
3  [2025-07-16T19:42:51.950Z]
   ----------------------------------------------------------------------
4  [2025-07-16T19:43:04.037Z] Iteration | Overflow |    HPWL (um)  | HPWL(%) |
   Penalty | Group
5  [2025-07-16T19:43:04.037Z]
   ------------------------------------------------------------
6  [2025-07-16T19:43:04.037Z]       1 | 0.9474 | 4.326570e+05 |  +0.00% | 4.77e
   -12 |
7  [2025-07-16T19:43:04.037Z]      10 | 0.9807 | 1.449570e+05 | -66.50% | 7.41e
   -12 |
8  [2025-07-16T19:43:04.037Z]      20 | 0.9807 | 1.404810e+05 |  -3.09% | 1.21e
   -11 |   ...
9  [2025-07-16T19:43:04.568Z] 3_1_place_gp_skip_io           2          186
   ed59792fd44b5db246b7...
10 [2025-07-16T19:43:07.240Z] 3_2_place_iop                  2          185 2079
   a626673768835cdf
11 [2025-07-16T19:43:07.525Z] /tmp/workspace/nROAD-flow-scripts-Public_master/flow/
   scripts/flow.sh 3_3_place_gp global_place
12 ...
13 [2025-07-16T19:43:10.068Z] Iteration | Overflow |    HPWL (um)  | HPWL(%) |
   Penalty | Group
14 [2025-07-16T19:43:10.068Z]
   ------------------------------------------------------------
15 [2025-07-16T19:43:10.068Z]       1 | 0.6664 | 8.177750e+05 |  +0.00% | 4.66e
   -11 |
16 [2025-07-16T19:43:10.068Z] [INFO GPL-0100] Timing-driven iteration 1/5, virtual:
   false.
17 [2025-07-16T19:43:10.068Z] [INFO GPL-0101]    Iter: 1, overflow: 0.666, keep resizer
   changes at: 1, HPWL: 817775
18 [2025-07-16T19:43:10.390Z] Iteration |   Area   | Resized | Buffers | Nets repaired
   | Remaining
19 [2025-07-16T19:43:10.390Z]
   ------------------------------------------------------------------
20 [2025-07-16T19:43:10.390Z]       0 |  +0.0% |    0 |    0 |              0
   |       621
21 [2025-07-16T19:43:10.390Z]   final |  +7.3% |  166 |    0 |              0
   |         0
22 [2025-07-16T19:43:10.390Z]
   ------------------------------------------------------------------
23 [2025-07-16T19:43:10.390Z] [INFO RSZ-0039] Resized 166 instances.
24 [2025-07-16T19:43:10.720Z]    Iter  |   Area  | Removed | Inserted | Pins
25 [2025-07-16T19:43:10.720Z]          |         | Buffers | Buffers  | Remaining
26 [2025-07-16T19:43:10.720Z] --------------------------------------------------
27 [2025-07-16T19:43:10.720Z]      0 |  +0.0% |    0 |    0 |       592
28 [2025-07-16T19:43:10.720Z]     59 |  +0.0% |    0 |    0 |       533
29 [2025-07-16T19:43:10.720Z]    118 |  +0.0% |    0 |    0 |       474
```

Figure 9: Segment of an open-source EDA log delineating its constituent as an example (OpenROAD Project 2025). For the main paper's experiments, we use the commercial EDA tool.



\end{document}